\documentclass[preprint,11pt]{elsarticle}

\usepackage{amssymb}
\usepackage{subfigure}

\usepackage[usenames]{color}
\usepackage{algorithmic}
\usepackage{algorithm}
\usepackage{cases}
\usepackage{array}
\usepackage{rotating}
\usepackage{longtable,lscape}
\usepackage[labelfont=bf]{caption} 
\usepackage{colortbl} 

\journal{Elsevier}

\begin{document}

\begin{frontmatter}



\title{A Multi-parent Memetic Algorithm for the Linear Ordering Problem}

\author[HUST]{Tao Ye}
\ead{yeetao@gmail.com}
\author[HUST]{Tao Wang}
\author[HUST]{Zhipeng L\"u}
\ead{zhipeng.lui@gmail.com}
\author[Angers]{Jin-Kao Hao}
\ead{hao@info.univ-angers.fr}
\address[HUST]{SMART, School of Computer Science and Technology, Huazhong University of Science and Technology, 430074 Wuhan, P.R.China}
\address[Angers]{LERIA, University of Angers, 2, Boulevard Lavoisier, 49045 Angers, France}

\renewcommand{\baselinestretch}{1.5}\large\normalsize
\begin{abstract}
\indent In this paper, we present a multi-parent memetic algorithm (denoted by MPM) for solving the classic Linear Ordering Problem (LOP). The MPM algorithm integrates in particular a multi-parent recombination operator for generating offspring solutions and a distance-and-quality based criterion for pool updating. Our MPM algorithm is assessed on 8 sets of 484 widely used LOP instances and compared with several state-of-the-art algorithms in the literature, showing the efficacy of the MPM algorithm. Specifically, for the 255 instances whose optimal solutions are unknown, the MPM is able to detect better solutions than the previous best-known ones for 66 instances, while matching the previous best-known results for 163 instances. Furthermore, some additional experiments are carried out to analyze the key elements and important parameters of MPM. 
\end{abstract}


\end{frontmatter}


\renewcommand{\baselinestretch}{1.5}\large\normalsize

\section{Introduction}
\label{introduction}
Given a $n\times n$ matrix $C$, the NP-hard Linear Ordering Problem (LOP) aims at finding a permutation $\pi $=$ (\pi_1,\pi_2,...,\pi_{n})$ of both the column and row indices $\{1,2,...,n\}$ which maximizes the following objective function:

\begin{equation}
\label{objectiv_function}
f({\pi})=\sum_{i=1}^{n}{\sum_{j=i+1}^{n}{C_{{\pi}_{i}{\pi}_{j}}}}
\end{equation}

In other words, the LOP is to identify a permutation of both the column and row indices of matrix $C$, such that the sum of the elements of the upper triangle (without the main diagonal) of the permuted matrix is maximized. This problem is equivalent to the maximum acyclic directed subgraph problem which, for a given digraph $G=(V,A)$ with arc weights $C_{ij}$ for each arc $(i,j) \in A$, is to find a subset $A' \subset A$ of arcs such that  $G=(V,A')$ is acyclic and $\sum_{(i,j) \in A'}{C_{ij}}$ is maximized \cite{Junger1996}.

The LOP has been the focus of numerous studies for a long time. It arises in a significant number of applications, such as the triangulation of input-output matrix in economy \cite{Leontief1986}, graph drawing \cite{Junger1996}, task scheduling \cite{Groschel1984}, determination of ancestry relationships \cite{Glover1974} and so on.

Due to its practical and theoretical importance, various solution algorithms have been proposed to solve the LOP. These algorithms can be divided into two main categories: exact algorithms and heuristic algorithms. Exact algorithms include, among others, a branch and bound algorithm \cite{Kass1981}, a branch and cut algorithm \cite{Groschel1984}, and a combined interior point/cutting plane algorithm \cite{Michell2000}. State-of-the-art exact algorithms can solve large instances from specific instance classes, but they may fail on other instances with much smaller size in the general case. Also, the computation time of exact algorithms may become prohibitive with the increase of the problem size.


The LOP is also tackled by a number of heuristic algorithms based on meta-heuristic approaches like {local search} \cite{Chanas1996}, elite tabu search \cite{Laguna1999}, scatter search \cite{Campos1999}, {iterated local search} \cite{Schiavinotto2004}, {greedy randomized adaptive search procedure} \cite{Campos2001}, {variable neighborhood search} \cite{Garcia2006} and the {memetic search} \cite{Schiavinotto2004}. In particular, according to the work of \cite{Marti2011}, the {memetic algorithm} of \cite{Schiavinotto2004} is the most successful among the state-of-the-art algorithms due to its excellent performance on the available LOP benchmark instances.

Inspired by the work of \cite{Schiavinotto2004}, this paper presents MPM, an improved memetic algorithm for solving the LOP. In addition to a local optimization procedure, the proposed MPM algorithm integrates two particular features. First, MPM employs a multi-parent recombination operator (denoted by \textsl{MPC}) to generate  offspring solutions which extends the {order based } (OB) operator \cite{Syswerda1991}. Second, MPM uses a distance-and-quality population updating strategy to keep a healthy diversity of the population.


We assess the MPM algorithm on 484 LOP instances widely used in the literature. For the 229 instances with known optimal solutions, the proposed algorithm can attain the optimal solutions consistently. For the remaining 255 instances whose optimal solutions are unknown, our algorithm is able to match the best-known results for 163 instances and in particular to find new solutions better than the previously best-known ones for 66 instances.


The remainder of this paper is structured as follows. Section \ref{Sec_Algorithms} presents in detail the MPM algorithm. Section \ref{Sec_Results} shows the computational statistics of MPM and comparisons with state-of-the-art algorithms. We will analyze some key elements and important parameters of MPM in Section \ref{Sec_Analysis}. 

\section{Multi-Parent Memetic Algorithm}
\label{Sec_Algorithms}

\subsection{Main Scheme}
\label{subsec_main_scheme}

The proposed MPM algorithm is based on the general memetic framework which combines the population-based evolutionary search and local search \cite{HandbookMA2012,Moscato2003} and follows the practical considerations  for discrete optimization suggested in \cite{Hao2012}. It aims at taking advantages of both recombination that discovers unexplored promising regions of the search space, and local search that finds good solutions by concentrating the search around these regions. 

The general MPM procedure is summarized in Algorithm \ref{Algo_mpha}. It is composed of four main components: a population initialing procedure, a local search procedure (Section \ref{subsec_LS}), a recombination operator (Section \ref{Recombination Operator}) and a population updating strategy  (Section \ref{subsec_pop_up}). Starting from an initial population of local optima obtained with the local search procedure, MPM performs a series of generations. At each generation, two or more solutions (parents) are selected in the population (Section \ref{MPC}) and recombined to generate an offspring solution (Section \ref{Recombination Operator}) which is improved by the local search procedure. The population is then updated with the improved offspring solution according to a distance-and-quality rule. In case the average solution quality of the population stagnates for $g$ generations, a new population is generated by making sure that the best solution found so far is always retained in the new population. This process continues until a stop condition is verified, such as a time limit or a fixed number of generation (Section \ref{subsec_reference_alg}).



\begin{algorithm}[H]
 \begin{small}
 \caption{Pseudo-code of the MPM algorithm}
 \label{Algo_mpha}
 \begin{algorithmic}[1]
 \STATE \textbf{INPUT:} matrix $C$, population size $p$, offspring size $c$
 \STATE \textbf{OUTPUT:} The best solution $s^*$ found so far
 \STATE $P = \{s^{1},s^{2},...,s^{p}\}$ $\leftarrow$ randomly generate $p$ initial solutions
 \FOR{$i = 1,2,\dots, p$}
 \STATE $s^i \leftarrow$  Local\_Search($s^i$) /* Section \ref{subsec_LS} */
 \ENDFOR

 \REPEAT
    \STATE Offspring $O\leftarrow\{\}$
    \FOR {$i = 1,2,\dots, c$}
        \STATE Choose $m$ individuals \{$s^{i1},...,s^{im}$\} from $P$ ($2\leq m \leq p$)/*Section \ref{MPC} */
        \STATE $s^o\leftarrow$ Recombination($s^{i1},...,s^{im}$)   /* Section \ref{Recombination Operator} */
        \STATE $s^o \leftarrow$ Local\_Search($s^o$)
        \STATE $O \leftarrow O\cup \{s^{o}\}$
    \ENDFOR
    \STATE $P \leftarrow$ Pool\_Updating($P,  O$)   /* Section \ref{subsec_pop_up} */
    \IF {Average solution quality stays the same for $g$ generations}
        \STATE Maintain the overall best solution $s^*$ in $P$
        \FOR {$i = 2,...,p$}
            \STATE Randomly generate an initial solution $s^{i}$
            \STATE $s^{i} \leftarrow$ Local\_Search($s^{i}$)
            \STATE $P \leftarrow P\cup \{s^{i}\}$
        \ENDFOR
    \ENDIF
 \UNTIL{termination condition is satisfied}
 \end{algorithmic}
 \end{small}
\end{algorithm}


\subsection{Local Search Procedure}
\label{subsec_LS}

Our local search procedure uses the neighborhood defined by the \textsl{insert} move which is very popular for permutation problems. An \textsl{insert} move is to displace an element in position $i$ to another position $j$ $(i \neq j)$ in the permutation sequence $\pi $=$ (\pi_1,\pi_2,...,\pi_{n})$.
\begin{equation}
insert(\pi,i,j) = \left\{
\begin{array}{r}
(...,\pi_{i-1},\pi_{i+1},...,\pi_j,\pi_i,\pi_{j+1},...), i<j
\\
(...,\pi_{j-1},\pi_{i},\pi_{j},...,\pi_{i-1},\pi_{i+1},...), i>j
\\
\end{array}
\right.
\end{equation}

It is clear that the size of this neighborhood is $(n-1)^2$.

To evaluate the neighborhood induced by the \textsl{insert} move, we introduce the $\Delta$-function, which indicates the changes in the objective function value caused by an \textsl{insert} move.

\begin{equation}
\Delta(\pi,i,j) = f(insert(\pi,i,j))-f(\pi)
\end{equation}

By using a fast evaluation method suggested in \cite{Congram2000}, the whole neighborhood can be examined with a time complexity of $O(n^2)$. More details about this evaluation method are given in \cite{Schiavinotto2004}).

Given a permutation $\pi$, our local search procedure selects at each iteration the best \textsl{insert} move (i.e., having the highest $\Delta$-value) to make the transition. This process repeats until we cannot find any \textsl{insert} move with a $\Delta$-value greater than zero. In this case, a local optimum is reached.

\subsection{Recombination Operator}
\label{Recombination Operator}

The recombination operator, which generates offspring solutions by combining features from parent individuals, is a relevant element in a memetic algorithm. In \cite{Schiavinotto2004}, four types of recombination operators (\textsl{Distance Preserving Crossover - DPX}, \textsl{Cycle Crossover - CX}, \textsl{Order-Based Crossover - OB} and \textsl{ Rank Crossover - Rank}) were compared for the LOP. According to the experiments, the \textsl{OB} operator \cite{Syswerda1991} performs the best among these four operators for the LOP. The general idea of the \textsl{OB} operator is to regard the permutation as a sequence and the operator tries to transmit the parent individuals' relative order to the offspring solution.

In this paper, we propose a newly designed adaptive multi-parent recombination operator (denoted by \textsl{MPC}) which can be considered as an extension of \textsl{OB}. The main difference between these two operators is that \textsl{MPC} uses three or more parent individuals to generate an offspring individual while \textsl{OB} is based on two parent individuals. As shown in Section \ref{Sec_Analysis}, this difference has a significant influence on the performance of the algorithm.

\subsubsection{General Ideas}
\label{subsubsec_GI}

In the LOP, a feasible solution is a permutation of $n$ elements and the good properties lie in the relative order of the $n$ elements imposed by the permutation. If we transmit the relative order in parent individuals to the offspring solution, the new solution keeps these elite components of their parents. Both \textsl{MPC} and \textsl{OB} operators are based on this basic idea.

\subsubsection{Parent Selection}
\label{subsubsec_PS}

Different from random parent selection technique used in \cite{Schiavinotto2004}, we employ a parent selection strategy which takes into consideration the distance between the selected solutions. Precisely, the proposed strategy relies on the notion of diversity of a population $P$ of solutions:


\begin{equation}
diversity(P)=\frac{\sum_{i=1}^{p-1}{\sum_{j=i+1}^{p}{dis(s^i,s^j)}}}{p\ast(p-1)/2}
\end{equation}

\noindent where $dis(s^i,s^j)$ is the distance between two solutions $s^i$ and $s^j$ defined as $n$ (the permutation length) minus the length of the longest common subsequence between $s^i$ and $s^j$ (also see Section \ref{subsec_pop_up}). Therefore, the population diversity takes values in [$0,n$].  

Our parent selection strategy determines a subset $SS$ of $m$ individuals from the population $P$ = $\{s^1,...,s^p\}$ such that the minimum distance between any two solutions in $SS$ is no smaller than a threshold:
\begin{equation}
\label{equation_in_ss}
min\{dis(s^i,s^j)|s^i,s^j\in SS\} \geq \beta*diversity(P)
\end{equation}

\noindent where $\beta \in [0,1]$ is a weighting coefficient which is fixed experimentally. Specifically, the subset $SS$ is constructed as follows. At each iteration, a solution is randomly selected from population $P$ and added into the subset $SS$ if Eq.\ref{equation_in_ss} is satisfied. Whenever such a solution exists, this process is repeated until subset $SS$ is filled with $m$ solutions. Otherwise, we reconstruct the subset $SS$ from scratch.

\subsubsection{Multi-parent Recombination Operator}
\label{MPC}

Now we describe how our \textsl{MPC} operator works to generate new offspring solutions. Recall that the conventional \textsl{OB} crossover uses two phases to generate an offspring individual $s^o$ from two parent individuals $s^1$ and $s^2$. In the first phase, $s^1$ is copied to $s^o$. In the second phase, \textsl{OB} selects $k$ (here, $k = n/2$) positions and reorders the elements in these $k$ selected positions according to their order in $s^2$. Readers are referred to \cite{Syswerda1991} for details of the \textsl{OB} operator.

Our \textsl{MPC} generalizes OB by employing $m$ ($m > 2$) parent individuals to generate a new offspring solution. Given $m$ selected parents \{$s^1,s^2,...,s^m$\}, the procedure of \textsl{MPC} also operates in two main phases. In the first phase, $s^1$ is copied to $s^o$. In the second phase, we repeatedly choose $k$ ($k=n/m$) different positions in $s^o$ and rearrange the elements in these chosen positions according to their order in $s^{i}$ $(2\leq i \leq m)$.

\begin{figure}[htbp]
\centering
\includegraphics[width=1\textwidth]{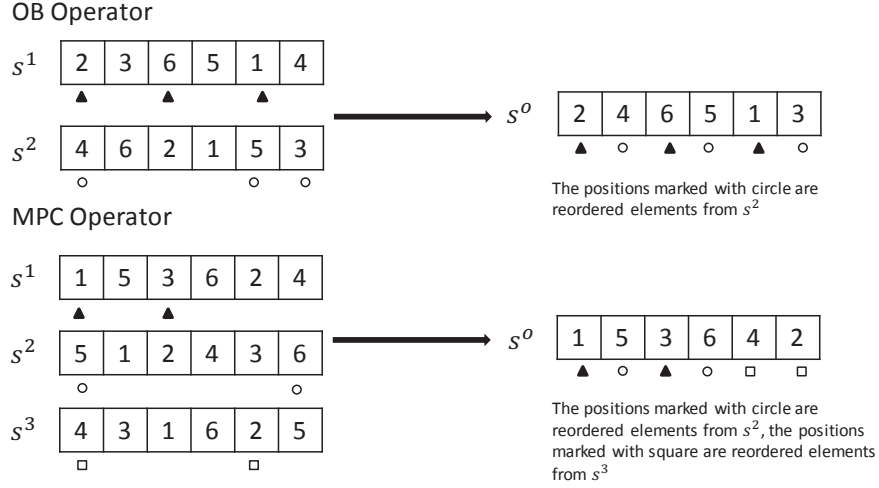}
\caption{An example of \textsl{MPC} and \textsl{OB}}
\label{Fig_sample}
\end{figure}

Fig.\ref{Fig_sample} shows an example of generating an offspring solution with \textsl{OB} and \textsl{MPC}. In this example, $n$ = $6$.
In the example of \textsl{OB}, $s^1$ is copied to $s^o$ first. Then we randomly choose positions ($2,4,6$) in $s^o$. The elements in these selected positions are ($3,5,4$) and these elements' relative order in $s^2$ is ($4,5,3$). So, we rearrange the selected elements according to their relative order in $s^2$.

In the example of \textsl{MPC}, $m$ = $3$ and $k$ = $2$. We generate $s^o$ in a similar way as \textsl{OB}. In the first step, $s^1$ is copied to $s^o$. In the second step, we randomly choose positions \textbf{\textsl{(2,4)}} in $s^o$ and rearrange the corresponding elements \textbf{\textsl{(5,6)}} according to their relative order in $s^2$, and then we randomly choose the positions \textbf{\textsl{(5,6)}} in  $s^o$ and rearrange the elements \textbf{\textsl{(2,4)}} according to their relative order in $s^3$.

\subsection{Pool Updating}
\label{subsec_pop_up}


In MPM, when $c$ offspring individuals have been generated by the multi-parent recombination operator, we immediately improve each of the offspring individuals with the local search procedure. Then we update the population with these improved offspring individuals. For the purpose of maintaining a healthy population diversity \cite{LuHao2010,Porumbeletal2010,SorensenSevaux2006}, we devise a distance-and-quality based population updating strategy. The idea is that if an offspring solution is not good enough or too close to the individuals in the population, it should not be added to the population. In our updating strategy, we first create a temporary population of size $p+c$ which is the union of the current population and the $c$ offspring individuals. Then we calculate for each individual $s$ a ``score'' by considering its quality and its distance to the other $p+c-1$ individuals. Finally, we choose the $p$-best individuals to form the new population. The notion of score is defined as follows.

\textbf{Definition 1}: (Distance between two solutions). Given two solutions $s^a$ = $(a_1,a_2,...,a_n)$ and $s^b$ = $(b_1,b_2,...,b_n)$, we define the distance between $s^a$ and $s^b$ as $n$ minus the length of their longest common subsequence (denoted by $LCS$).
\begin{equation}
dis(s^a,s^b) = n-LCS(s^a,s^b)
\end{equation}

It is clear that a small value of $dis(s^a,s^b)$ indicates that the two solutions are similar to each other. The time complexity of calculating this distance is $O(n^2)$ \cite{Cormen2001}.

\textbf{Definition 2}: (Distance between one solution and a population). Given a solution $s^a$ = $(a_1,a_2,...,a_n)$ and a population $P$ = $\{s^1,s^2,...,s^p\}$, the distance between $s^a$ and $P$ is the minimum distance between $s^a$ and $s^i$ $(1 \leq i \leq p)$.
\begin{equation}
dis(s^a,P) = min\{dis(s^a,s^i),(1\leq i \leq p),s^a\neq s^i\}
\label{dis_p_sol}
\end{equation}

\textbf{Definition 3}: (Score of a solution with respect to a population). Given a population $P$ = $\{s^1,s^2,...,s^p\}$, the score of a solution $s^i$ in $P$ is defined as
\begin{equation}
\label{eq_score_in_p}
score(s^i,P)= \alpha\widetilde{A}(f(s^i))+(1-\alpha)\widetilde{A}(dis(s^i,P))
\end{equation}
\noindent where $f(s^i)$ is the objective function value of solution $s^i$, and $\widetilde{A}()$ represents the normalized function:

\begin{equation}
\widetilde{A}(y)=\frac{y-y_{min}}{y_{max}-y_{min}+1}
\label{equ_socre}
\end{equation}
\noindent where $y_{max}$ and $y_{min}$ are respectively the maximum and minimum values of $y$ in $P$. The number $1$ is added to avoid $0$ denominator. $\alpha$ is a parameter to balance the two parts of quality and distance.

The score function is thus composed of two parts. The first part concerns the quality (objective function value) while the second part considers the diversity of the population. It is easy to check that if a solution has a high score, it is of good quality and is not too close to the other individuals in the population. Algorithm \ref{Algo_up} describes the pseudo-code of our pool updating strategy.  

\begin{algorithm}[H]
 \begin{small}
 \caption{Pseudo-code of population updating}
 \label{Algo_up}
 \begin{algorithmic}[1]
 \STATE \textbf{INPUT:} Population $P$ = $\{s^1,...,s^p\}$ and Offspring $O$ = $\{o^1,...,o^c\}$
 \STATE \textbf{OUTPUT:} Updated population $P$ = $\{s^1,...,s^p\}$
 \STATE $P$ : $P' \leftarrow P \cup O$   /* Tentatively add all offspring $O$ to population $P$ */
 \FOR{$i = 1,...,p+c$}
 \STATE Calculate the distance between $s^i$ and $P$ according to Eq. \ref{dis_p_sol}
 \ENDFOR
 \FOR{$i = 1,...,p+c$}
 \STATE Calculate the score of each $s^i$ in $P$ according to Eq. \ref{eq_score_in_p}
 \ENDFOR
 \STATE Sort the individuals in non-decreasing order of their scores
 \STATE Choose the $p$ best individuals to form $P$
 \RETURN $P$
 \end{algorithmic}
 \end{small}
\end{algorithm}


\section{Computational Results and Comparisons}
\label{Sec_Results}

In this section, we report experimental evaluations of our MPM algorithm by using the well-known LOLIB benchmark instances. We show computational results and compare them with the best known results obtained by the state-of-the-art algorithms in the literature.

\subsection{Problem Instances and Experimental Protocol}
\label{subsec_reference_alg}

The LOLIB benchmarks have 484 instances in total and they are divided into 8 sets\footnote{All the instances are available at: http://www.optsicom.es/lolib/}. The optimal solutions and best-known results for each instance can be found in \cite{Marti2011}.
\\ \indent \textbf{IO}: This is a well-known set of instances that contains 50 real-world linear ordering problems generated from input-output tables from various sources. It was first used in \cite{Groschel1984}.
\\ \indent \textbf{SGB}: These instances are from \cite{Knuth} and consist of input-output tables from sectors of the economy of the United States. The set has a total of 25 instances with 75 sectors.
\\ \indent \textbf{RandAI}: There are 25 instances in each set with $n$ = 100, 150, 200 and 500, respectively, giving a total of 100 instances.
\\ \indent \textbf{RandAII}: There are 25 instances in each set with $n$ = 100, 150 and 200, respectively, giving a total of 75 instances.
\\ \indent \textbf{RandB}: 90 more random instances.
\\ \indent \textbf{MB}: These instances have been used by Mitchell and Borchers for their computational experiments.
\\ \indent \textbf{xLOLIB}: Some further benchmark instances have been created and used by Schiavinotto and St\"{u}tzle \cite{Schiavinotto2004}, giving a total of 78 instances.
\\ \indent \textbf{Special}: 36 more instances used in \cite{Christof1997,Christof1996,Goemans1996}.

\renewcommand{\baselinestretch}{1.0}\large\normalsize
\begin{table}[!ht]\centering \scriptsize
\begin{center}
\captionsetup{font={small}}
\caption{Sets of the tested instances}
\label{table_sum_instance}
\begin{tabular}{|l|c|c|cc|}
\hline
& & & & \\
\centering{\raisebox{1.60ex}[0cm][0cm]{Set}}& \centering{\raisebox{1.60ex}[0cm][0cm]{\#Instances}}& \centering{\raisebox{1.60ex}[0cm][0cm]{\#Optimal}}& \centering{\raisebox{1.60ex}[0cm][0cm]{\#Lower Bound}}& \\
\hline
\centering{\textbf{IO}} & \centering{50} & \centering{50} & \centering{-} & \\
\centering{\textbf{SGB}} & \centering{25} & \centering{25} & \centering{-} & \\
\centering{\textbf{RandAI}} & \centering{100} & \centering{-} & \centering{100} & \\
\centering{\textbf{RandAII}} & \centering{75} & \centering{25} & \centering{50} & \\
\centering{\textbf{RandB}} & \centering{90} & \centering{70} & \centering{20} & \\
\centering{\textbf{MB}} & \centering{30} & \centering{30} & \centering{-} & \\
\centering{\textbf{xLOLIB}} & \centering{78} & \centering{-} & \centering{78} & \\
\centering{\textbf{Special}} & \centering{36} & \centering{29} & \centering{7} & \\
\hline
\centering{\textbf{Total}} & \centering{484} & \centering{229} & \centering{255} & \\
\hline
\end{tabular}
\end{center}
\end{table}
\renewcommand{\baselinestretch}{1.5}\large\normalsize

Table \ref{table_sum_instance} summarizes the number of instances in each instance class described above together with the information about the number of instances whose optimal solutions or lower bounds are known.

Our MPM algorithm is programmed in C and compiled using GNU GCC on a PC running Windows XP with 2.4GHz CPU and 2.0Gb RAM. Given the stochastic nature of the MPM, we solved each problem instance independently 50 times using different random seeds subject to a time limit of 2 hours. Note that the best known results listed in the following tables are also obtained within 2 hours, which are available at: http://www.optsicom.es/lolib/.

\subsection{Parameter Setting}
\label{subsec_ps}

Like all previous heuristic algorithms, MPM uses several parameters which are fixed via a preliminary experiments with a selection of problem instances. Precisely, we set $p$ = 25, $c$ = 10 and $g$ = 30 ($p$ is the population size, $c$ is the offspring size and $g$ means that if the average solution quality stays unchanged for $g$ generations, the population is reconstructed). In the light of the experiments carried out in Section \ref{subsec_Analysis_mpc}, we choose $m$ = 3 to be the number of parents, $\beta$ = $rand(0.6,0.7)$ for the parent selection and $\alpha$ = $rand(0.8,1.0)$ for the pool updating strategy. These settings are used to solve all the instances without any further fine-tuning of the parameters.

\subsection{Computational Results}
\label{subsec_results}

We aim to evaluate the MPM's performance on the LOLIB benchmark instances, by comparing its performance with the best-known results in the literature. Table \ref{table_res_opt_instance} summarizes the computational statistics of our MPM algorithm on the instances with known optimal solutions. The name of each instance set is given in column 1, column 2 shows the number of instances for which the optimal results are obtained, column 3 gives the number of instances for which our algorithm matches the optimal solutions, column 4 presents the deviation from the optimal solutions and the average CPU time to match the optimal solutions is given in column 5.

\renewcommand{\baselinestretch}{1.0}\large\normalsize
\begin{table}[!ht]\centering \scriptsize
\begin{center}
\captionsetup{font={small}}
\caption{MPM's performance on the instances with known optimal solutions}
\label{table_res_opt_instance}
\begin{tabular}{|l|c|c|c|cc|}
\hline
& & & & & \\
\centering{\raisebox{1.60ex}[0cm][0cm]{Set}}& \centering{\raisebox{1.60ex}[0cm][0cm]{\#Optimal}}& \centering{\raisebox{1.60ex}[0cm][0cm]{\#Match Optimal}}& \centering{\raisebox{1.60ex}[0cm][0cm]{\#Dev(\%)}}&
\centering{\raisebox{1.60ex}[0cm][0cm]{Time(s)}}& \\
\hline
\centering{\textbf{IO}} & \centering{50} & \centering{50} & \centering{0.0} & \centering{$<$1} & \\
\centering{\textbf{SGB}} & \centering{25} & \centering{25} & \centering{0.0} & \centering{$<$1} & \\
\centering{\textbf{RandAII}} & \centering{25} & \centering{25} & \centering{0.0} & \centering{$<$10} & \\
\centering{\textbf{RandB}} & \centering{70} & \centering{70} & \centering{0.0} & \centering{$<$10} & \\
\centering{\textbf{MB}} & \centering{30} & \centering{30} & \centering{0.0} & \centering{$<$1} & \\
\centering{\textbf{Spec}} & \centering{29} & \centering{29} & \centering{0.0} & \centering{$<$10} & \\
\hline
\centering{\textbf{Total}} & \centering{229} & \centering{229} & \centering{0.0} & \centering{-} & \\
\hline
\end{tabular}
\end{center}
\end{table}
\renewcommand{\baselinestretch}{1.5}\large\normalsize

\renewcommand{\baselinestretch}{1.0}\large\normalsize
\begin{table}[!ht]\centering \scriptsize
\begin{center}
\captionsetup{font={small}}
\caption{MPM's performance on the instances with best-known solutions}
\label{table_res_ub_instance}
\begin{tabular}{|l|c|c|c|cc|}
\hline
& & & & & \\
\centering{\raisebox{1.60ex}[0cm][0cm]{Set}}& \centering{\raisebox{1.60ex}[0cm][0cm]{\#Lower Bound}}& \centering{\raisebox{1.60ex}[0cm][0cm]{\#Match Best-known}}& \centering{\raisebox{1.60ex}[0cm][0cm]{\#Improve Best-known}}&
\centering{\raisebox{1.60ex}[0cm][0cm]{Dev.B(\%)}}& \\
\hline
\centering{\textbf{RandAI}} & \centering{100} & \centering{64} & \centering{33} & \centering{0.015} & \\
\centering{\textbf{RandAII}} & \centering{50} & \centering{50} & \centering{0} & \centering{0.0} & \\
\centering{\textbf{RandB}} & \centering{20} & \centering{20} & \centering{0} & \centering{0.0} & \\
\centering{\textbf{xLOLIB}} & \centering{78} & \centering{24} & \centering{32} & \centering{0.046} & \\
\centering{\textbf{Spec}} & \centering{7} & \centering{5} & \centering{1} & \centering{0.2} & \\
\hline
\centering{\textbf{Total}} & \centering{255} & \centering{163} & \centering{66} & \centering{0.02} & \\
\hline
\end{tabular}
\end{center}
\end{table}
\renewcommand{\baselinestretch}{1.5}\large\normalsize

\renewcommand{\baselinestretch}{1.0}\large\normalsize
\begin{center}
\scriptsize
\begin{longtable}{|c|c|c|}
\captionsetup{font={small}}
\caption{Detailed results on the 66 instances for which MPM can improve the previous best-known results}
\label{table_res_improve_instance} \\
\hline \multicolumn{1}{|c|}{\textbf{Instance}} & \multicolumn{1}{c|}{\textbf{Bound[LB,UB]}} & \multicolumn{1}{c|}{\textbf{Our Results}} \\ \hline
\endfirsthead
\multicolumn{3}{c}%
{{\bfseries \tablename\ \thetable{} -- Continued from previous page}} \\
\hline \multicolumn{1}{|c|}{\textbf{Instance}} &
\multicolumn{1}{c|}{\textbf{Bound[LB,UB]}} &
\multicolumn{1}{c|}{\textbf{Our Results}} \\ \hline
\endhead
\hline \multicolumn{3}{|r|}{{Continued on next page}} \\ \hline
\endfoot
\hline \hline
\endlastfoot
{N-atp163}  &  [2073,2417] &  2074 \\
{N-be75np\_150}   &  [7174395,7317546] &  7174972  \\
{N-be75np\_250}  & [17819028,18473322] & 17819139 \\
{N-be75oi\_150}  &  [2246534,2259482] &  2246571  \\
{N-be75oi\_250}   &  [5910266,5978555] &  5910492  \\
{N-be75tot\_150}   &  [12287707,12509023]  &  12288645  \\
{N-be75tot\_250}   &  [30993002,32055676]  &  30993138  \\
{N-stabu1\_250}   &  [7744014,8012535]  &  7744106  \\
{N-stabu2\_150}   &  [4327538,4398662]  &  4328230  \\
{N-t59d11xx\_250}   &  [3842563,4015773]  &  3843449  \\
{N-t65f11xx\_150}   &  [3159526,3231148]  &  3159539  \\
{N-t65l11xx\_250}   &  [666664,679527]  &  666683  \\
{N-t65n11xx\_150}   &  [550849,558953]  &  550856  \\
{N-t69r11xx\_250}   &  [31824632,32688871]  &  31824787  \\
{N-t70b11xx\_150}   &  [9649306,9802850]  &  9649316  \\
{N-t70b11xx\_250}   &  [25411146,26133557]  &  25411943  \\
{N-t70d11xn\_150}   &  [5825509,5956793]  &  5825719  \\
{N-t70d11xn\_250}   &  [15212874,15833337]  &  15215721  \\
{N-t70l11xx\_150}   &  [436862,438087]  &  436863  \\
{N-t74d11xx\_250}   &  [24444287,25350257]  &  24445713  \\
{N-t75d11xx\_150}   &  [9644779,9850135]  &  9645000  \\
{N-t75k11xx\_250}   &  [4094877,4249417]  &  4094905  \\
{N-tiw56n54\_250}   &  [2099740,2182012]  &  2099742  \\
{N-tiw56n58\_250}   &  [2906872,3026780]  &  2907355  \\
{N-tiw56n62\_150}   &  [1626927,1668859]  &  1626966  \\
{N-tiw56n62\_250}   &  [4143260, 4317513]  &  4145133 \\
{N-tiw56n66\_250}   &  [5371361,5582715]  &  5371522  \\
{N-tiw56n67\_150}   &  [2372926,2422085]  &  2372945  \\
{N-tiw56n72\_150}   &  [4135907,4222250]  &  4135952  \\
{N-tiw56r54\_150}   &  [958192,979551]  &  958195  \\
{N-tiw56r58\_250}   &  [3060323,3185528]  &  3060360  \\
{N-tiw56r66\_250}   &  [4948720,5417594]  &  4949345  \\
{N-tiw56r67\_250}   &  [5292028,5491577]  &  5292409  \\
{N-t1d200.02}   &  [407729,461223]  &  407733  \\
{N-t1d200.04}   &  [410101, 465271]  &  410120  \\
{N-t1d200.08}   &  [408850, 462476]  &  408857  \\
{N-t1d200.13}   &  [409234, 459659]  &  409270  \\
{N-t1d200.18}   &  [407709, 467958]  &  407822  \\
{N-t1d200.20}   &  [406418, 455487]  &  406420  \\
{N-t1d200.22}   &  [407333, 458596]  &  407377  \\
{N-t1d200.25}   &  [406356,458197]  &  406476  \\
{N-t1d500.1}   &  [2402774,4191813]  &  2404108  \\
{N-t1d500.2}   &  [2411570,4207198]  &  2412011  \\
{N-t1d500.3}   &  [2404784,4205918]  &  2404815  \\
{N-t1d500.4}   &  [2413600,4221950]  &  2414671  \\
{N-t1d500.5}   &  [2391486,4186810]  &  2392298  \\
{N-t1d500.6}   &  [2399394,4190956]  &  2401386  \\
{N-t1d500.7}   &  [2400739,4198457]  &  2400740 \\
{N-t1d500.8}   &  [2413108,4206654]  &  2414166  \\
{N-t1d500.9}   &  [2406343,4198840]  &  2407173  \\
{N-t1d500.10}   &  [2404420, 4198760]  &  2405923  \\
{N-t1d500.11}   &  [2416364,4210737]  &  2416813  \\
{N-t1d500.12}   &  [2402581,4194185]  &  2403302  \\
{N-t1d500.13}   &  [2405118,4197442]  &  2406446   \\
{N-t1d500.14}   &  [2410693,4200887]  &  2410694   \\
{N-t1d500.15}   &  [2411718,4208905]  &  2412599   \\
{N-t1d500.16}   &  [2416067,4200206]  &  2416346   \\
{N-t1d500.17}   &  [2401800,4197344]  &  2402784   \\
{N-t1d500.18}   &  [2421159,4222286]  &  2422227   \\
{N-t1d500.19}   &  [2404029,4198658]  &  2404236   \\
{N-t1d500.20}   &  [2414713,4207789]  &  2415218   \\
{N-t1d500.21}   &  [2405615,4201350]  &  2406326   \\
{N-t1d500.22}   &  [2408164,4208557]  &  2409413   \\
{N-t1d500.23}   &  [2408689,4197731]  &  2408042   \\
{N-t1d500.24}   &  [2402712,4191909]  &  2403229   \\
{N-t1d500.25}   &  [2405718,4196590]  &  2405990   \\
\end{longtable}
\end{center}
\renewcommand{\baselinestretch}{1.5}\large\normalsize

Table \ref{table_res_ub_instance} shows the performance of our MPM algorithm on the instances whose optimal solutions are unknown. In Table \ref{table_res_ub_instance}, the name of each set is presented in column 1. Column 2 gives the number of instances for which the best-known results are achieved. Columns 3-4 present the number of instances for which we match and improve the previous best-known results, respectively. The deviation from the best-known results is presented in column 5. Table \ref{table_res_improve_instance} describes the details of the instances for which our algorithm can improve the previous best-known solutions.

When compared with the best-known results reported in the literature, one observes that for the 229 instances with known optimal values, our MPM algorithm can match all the optimal results within 10 seconds. In addition, for the 255 instances whose optimal solutions are unknown, our MPM algorithm can match the previous best known results for 163 instances while improving the best known results for 66 ones. In sum, our MPM algorithm can match 392 previous best results and find improved solutions for 66 instances for the 484 instances. The results obtained by our MPM are thus quite competitive compared with the previous best-known results with respect to the solution quality. This experiment demonstrates the competitiveness of our proposed algorithm.


In \cite{Marti2011}, 10 best performing algorithms are compared with each other under a comparable computational time, which include: {KLM}-Kernighan and Lin multi-start method in \cite{Kernighan1979}, {CKM}-Chanas and Kobilansky multi-start method in \cite{Chanas1996},
{GRASP}-Greedy Randomized Adaptive Search Procedure in \cite{Campos2001}, {TS}-Tabu Search in \cite{Laguna1999}, {SS}-Scatter Search in \cite{Campos1999}, {VNS}-Variable Neighborhood Search in \cite{Garcia2006}, {GA} in \cite{Huang2003}, {MA} in \cite{Schiavinotto2004},
{SA}-Simulated Annealing in \cite{Marti2011} and {ILS}-Iterated Local Search in \cite{Schiavinotto2004}.

A summarized comparison among these 10 state-of-the-art algorithms is presented in \cite{Marti2011} separately for the instances with known and unknown optimal solutions, where the deviation to the best known results and the success rate for hitting the best known results are reported. According to this comparison, \textsl{MA} in \cite{Schiavinotto2004} shows the best performance for the LOP. In addition, \textsl{ILS} in \cite{Schiavinotto2004} is also very competitive
in terms of effectiveness and efficiency.

Although we cannot exactly compare our MPM with these reference algorithms due to the reason that different computational environments are employed and only a summarized results of these reference algorithms are given in \cite{Marti2011}. However, if we compare these summarized results of the reference algorithms with our results reported in Tables \ref{table_res_opt_instance} and \ref{table_res_ub_instance}, we can make the followings comments. First, for the instances with known optimal values, our MPM can achieve all the optimal solution within 10 seconds, which is quite competitive with the reference algorithms in terms of both the deviation to the best known results and the success rate. Second, for the instances with unknown optimal solutions, MPM matches all the previous best-known results while improving the previous best known results for 66 instances but with a longer CPU time. Third, if we use the same time limit as the reference algorithms, our MPM can still obtain competitive results with respect to both the deviation to the best known results and the success rate when comparing with the best performing algorithms MA and ILS of \cite{Schiavinotto2004}. From these observations, one finds that our MPM algorithm competes favorably with these best performing algorithms in the literature, especially on the instances that are considered to be challenging.



\section{Analysis and Discussion}
\label{Sec_Analysis}

We now turn our attention to discussing and analyzing two ingredients implemented in MPM, namely, the \textsl{MPC} recombination operator and the population updating strategy. 

\subsection{Comparison between MPC and OB}
\label{subsec_Analysis_mpc}


As indicated in Section \ref{MPC}, the \textsl{MPC} operator can be regarded as an extended version of the \textsl{OB} operator in \cite{Syswerda1991}. In order to ensure that this extension makes a meaningful contribution to our proposed algorithm, we carry out experiments to compare \textsl{MPC} with OB. Keeping other ingredients unchanged in our MPM algorithm, we compare the MPM algorithm with different $m$-parent recombination operators, where $m$ = $2,3,4$, respectively. Note that $m$ = $2$ corresponds to the original \textsl{OB} operator.

\begin{figure}[htbp]
\centering
\includegraphics[width=0.7\textwidth]{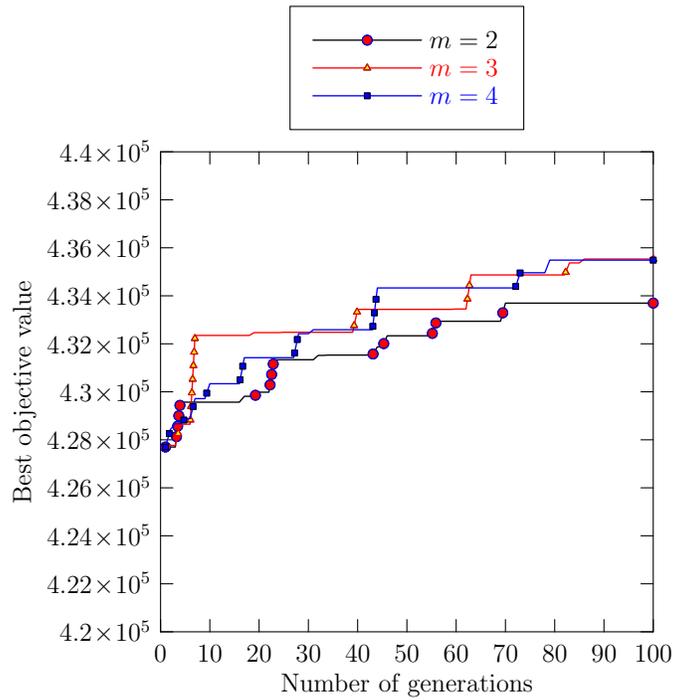}
\caption{Difference between \textsl{MPC} and \textsl{OB} in terms of solution quality}
\label{Fig_mpc_no_stra}
\end{figure}

In our experiments, we observe the best objective function value evolving with the number of generations. We compare the performance of \textsl{MPC} and \textsl{OB} by observing how the best objective function value evolves when the search progresses. As an illustration we show in Fig. \ref{Fig_mpc_no_stra} the results of this experiment based on the instance N-t70l11xx\_150 with a fixed number of 100 generations. Similar phenomenon can be observed on other instances. From Fig.\ref{Fig_mpc_no_stra}, one finds that \textsl{MPC} ($m = 3,4$) is more effective compared with the \textsl{OB} operator with respect to the objective value. But there is no obvious difference between the two multiple parent strategies ($m = 3,4$). 

In order to show that \textsl{MPC} is able to generate high quality solutions, we also conducted experiments on some challenging xLOLIB instances. In this experiment, the total number of generations is set to be 400 and our algorithm is independently run 10 times. Table \ref{table_additional_experiments} shows the best and average objective values for different \textsl{MPC} operators with $m=2,3$ and 4 respectively.

\renewcommand{\baselinestretch}{1.5}\small\normalsize
\begin{table}
\centering \scriptsize
\caption{Comparison between \textsl{MPC} with different number of parents}
\label{table_additional_experiments}
\begin{tabular}{p{1.7cm}|p{1.4cm}p{1.4cm}p{1.4cm}p{1.4cm}p{1.4cm}p{1.4cm}p{0.00cm}}
\hline
& \multicolumn{2}{c}{$m = 2$} & \multicolumn{2}{c}{$m = 3$} & \multicolumn{2}{c}{$m = 4$}\\
\cline{2-3}
\cline{4-5}
\cline{6-7}
\centering{\raisebox{2.50ex}[0cm][0cm]{Instance}} & \centering{$f_{best}$} & \centering{$f_{avg}$} & \centering{$f_{best}$} & \centering{$f_{avg}$} & \centering{$f_{best}$} & \centering{$f_{avg}$} &
\\
\hline
\centering{t59b11xx\_150} & \centering{8382725} & \centering{8377458.0} & \centering{\textbf{8387360}} & \centering{8377096}  & \centering{\textbf{8388760}} & \centering{\textbf{8379571.0}}  & \\
\centering{t59n11xx\_150} & \centering{318570} & \centering{318411.2} & \centering{318781} & \centering{\textbf{318505.0}}  & \centering{\textbf{318792}} & \centering{318452.4}  & \\
\centering{t65n11xx\_150} & \centering{549459} & \centering{\textbf{549355.0}} & \centering{\textbf{549679}} & \centering{548917.6}  & \centering{549391} & \centering{548708.8}  & \\
\centering{t70l11xx\_150} & \centering{436580} & \centering{436338.2} & \centering{436503} & \centering{\textbf{436456.8}}  & \centering{\textbf{436582}} & \centering{436368.2}  & \\
\centering{tiw56r54\_150} & \centering{957002} & \centering{\textbf{956600.2}} & \centering{957301} & \centering{956442.0}  & \centering{\textbf{957712}} & \centering{956529.6}  & \\
\centering{be75oi\_250} & \centering{5895283} & \centering{\textbf{5891278.2}} & \centering{5892923} & \centering{5888500.0}  & \centering{\textbf{5901134}} & \centering{5890684.8}  & \\
\centering{t70d11xx\_250} & \centering{15998903} & \centering{15982525} & \centering{16011961} & \centering{\textbf{16001681.0}}  & \centering{\textbf{16012749}} & \centering{15996008.8}  & \\
\centering{t75k11xx\_250} & \centering{4086690} & \centering{4083981.6} & \centering{\textbf{4094070}} & \centering{\textbf{4087744.8}}  & \centering{4083293} & \centering{4079735.4}  & \\
\centering{stabu1\_250} & \centering{7724593} & \centering{\textbf{7720036.4}} & \centering{\textbf{7728866}} & \centering{7719452.4}  & \centering{7728885} & \centering{7719805.8}  & \\
\centering{tiw56r67\_250} & \centering{5281958} & \centering{5279749.2} & \centering{\textbf{5282895}} & \centering{\textbf{5279979.3}}  & \centering{5281431} & \centering{5276231.2}  & \\

\hline
\end{tabular}
\end{table}
\renewcommand{\baselinestretch}{1.5}\large\normalsize

From Table \ref{table_additional_experiments}, one observes that for these challenging instances, \textsl{MPC} ($m$ = $3,4$) performs better than the \textsl{OB} operator in terms of the best objective function values ($f_{best}$). In terms of average objective value ($f_{avg}$), \textsl{OB} may outperform \textsl{MPC} on some instances, which shows that \textsl{OB} may have difficulties in finding the best solutions but it is more stable when compared with \textsl{MPC}. 

These results show that the traditional 2-parent recombination operator has an overall good performance in terms of $f_{avg}$, but it may fail to find the best solutions. The \textsl{MPC} operator, by contrast, favors in producing promising offspring solutions in many situations, though it is less successful in few cases.





\subsection{Population Updating Strategy}

\begin{figure}[htbp]
\centering
\includegraphics[width=1\textwidth]{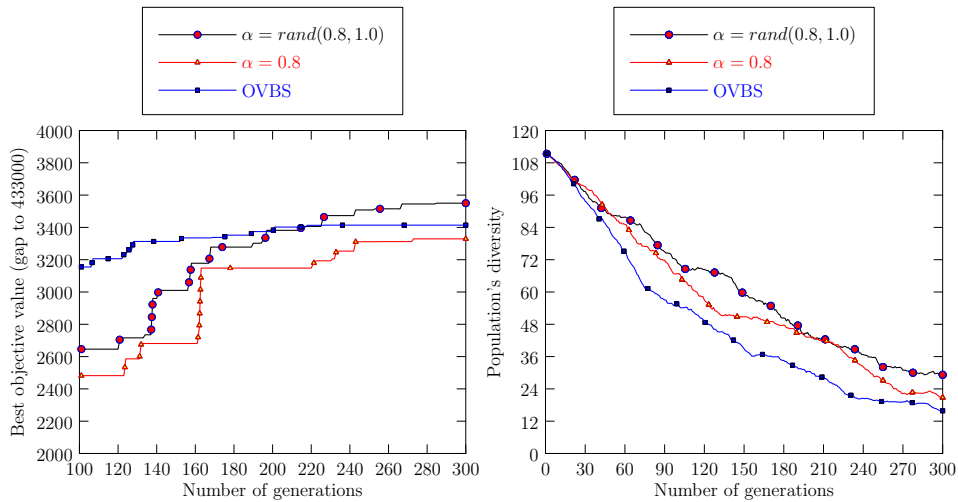}
\caption{Difference between strategies for pool updating}
\label{Fig_up with_stra}
\end{figure}

Our pool updating strategy uses a quality-and-distance scoring function to rank the individuals of the population. Experiments are carried out to verify how the parameter $\alpha$ in Eq. \ref{eq_score_in_p} affects the performance of our algorithm. We tested different $\alpha$ values, namely, $\alpha=0.8$, and random value between 0.8 and 1.0, denoted as $rand(0.8,1.0)$. We also compare our strategy with the strategy proposed in \cite{Schiavinotto2004}. Considering the fact that the strategy in \cite{Schiavinotto2004} only takes objective function value into account, we call it ``Only Value Based Strategy'' (denoted as \textsl{OVBS}). This experiment is illustrated in Fig. \ref{Fig_up with_stra} on the instance N-t70l11xx\_150.

From Fig.\ref{Fig_up with_stra}, we find that $rand(0.8,1.0)$ strategy performs better than the \textsl{OVBS} and $\alpha=0.8$ strategies in terms of the objective value. At the beginning of the search, the \textsl{OVBS} can find better solutions, but our algorithm can find better solutions at the end of the search when the number of generations reaches 300, i.e., $f=436550$ versus $f=436329$ for \textsl{OVBS} and versus $f=436414$ for $\alpha=0.8$. When it comes to population diversity, our $rand(0.8,1.0)$ strategy also makes a difference. As the algorithm processes, our strategy enhances the population diversity compared with the \textsl{OVBS} strategy.

\section{Conclusions}
\label{Sec_Conclusions}

In this paper, we have proposed MPM, a multi-parent memetic algorithm for solving the linear ordering problem. The proposed algorithm integrates several particular features, such as a multi-parent recombination operator, a diversity based parent selection strategy and a quality-and-diversity based pool updating strategy. Computational results on 8 sets of 484 popular LOP instances and comparisons with 10 reference algorithms in the literature demonstrate the efficacy of our algorithm. In particular, our MPM algorithm detects better lower bounds than the previous best ones for 66 challenging instances.

In addition, we carried out experiments to study two ingredients of MPM. Our experiments demonstrate the merit of the multi-parent recombination operator with respect to the traditional 2-parent operator. Furthermore, the computational results also show that the distance-and-quality based pool updating strategy provides the population with a good diversity. These features together lead to the observed high performance of our algorithm.


\section*{Acknowledgments}
This work was partially supported by the National Natural Science Foundation of China (Grant No. 61100144), the Doctoral Fund of Ministry of Education of China (Grant No. 20110142120081) and the RaDaPop (2009-2013) and LigeRo projects (2009-2013) from Pays de la Loire Region, France.

\end{document}